# Lexicons and grammars for language processing: industrial or handcrafted products?

Éric Laporte[1]

During the recent years, the use of linguistic data for language processing (semantic ambiguity resolution, translation...) increased progressively. Such data are now commonly called language resources. A few years ago, nearly all the language resources used for this purpose were collections of texts as the Brown Corpus and the Penn Treebank, but the use of electronic lexicons (WordNet, FrameNet, VerbNet, ComLex...) and formal grammars (TAG...) developed recently. This development is slow because of most processes of construction of lexicons and grammars are manual, whereas the construction of corpora has always been highly automated.

However, more and more specialists of language processing realize that the information content of lexicons and grammars is richer than that of corpora, and hence the former make more elaborate processing possible. The difference in construction time is likely to be connected with the difference in information content: the handcrafting of lexicons and grammars by linguists would make them more informative than automatically generated data.

This situation can evolve into two directions: either specialists of language technology get progressively used to handling manually constructed resources, which are more informative and more complex, or the process of construction of lexicons and grammars is automated and industrialized, which is the mainstream perspective. Both evolutions are already in progress, and a tension exists between them. The relation between linguists and computer scientists depends on the future of these evolutions, since the first implies training and hiring numerous linguists, whereas the other depends essentially on solutions elaborated by computer engineers.

The aim of this article[2] is to analyse practical examples of the language resources in question, and to discuss about which of the two trends, handcrafting or generating industrially, or a combination of both, can give the best results or is the most realistic.

This article is organized as follows. In the next section, we introduce basic notions about language resources, corpora, lexicons and grammars. In section 2, we provide technical facts about manually constructed lexicons and grammars. Section 3 analyses the situation of tension between the handcrafting approach and the industrial approach. Sections 4 and 5 discuss the points in favour of the respective trends. The article ends with conclusive remarks.

## 1. Basic notions

Language processing is a field that studies the computer processing and generation of written text and other linguistic productions. The best-known application of this field is information retrieval, the task perfomed by search engines. These systems work with practically no specific knowledge about the languages of the texts they process. However, there are many other applications, and some of them are much more complex, for instance semantic ambiguity resolution and translation. Language-processing application systems have progressively used more and more data about

[1] Université Paris-Est - Laboratoire d'informatique de l'Institut Gaspard-Monge (IGM-Labinfo) - 5, bd Descartes - F77454 Marne-la-Vallée CEDEX 2 - France - eric.laporte@univ-paris-est.fr

[2] We thank the organizers of the VIIth Seminar of Research of the Linguistics and Portuguese Language Postgraduation Program, in the Araraquara campus of the Universidade Estadual Paulista (UNESP), for their invitation to give a talk and to publish this article. We are also indebted to Duško Vitas, from the University of Belgrade, for his comments about a previous version of this article. Of course, he is not responsible for what we say.

languages since 1999. Such data are now commonly called 'language resources'. They assume three main forms:
- annotated corpora, i.e. collections of texts annotated with information attached to words or other parts of the texts,
- electronic lexicons or dictionaries, i.e. descriptions of properties of words,
- formal grammars, i.e. descriptions of rules of combination of words.

Since 1998, a series of biennial international conferences, the Language Resources and Evaluation Conference, is devoted to these resources.

### 1.1 Annotated corpora

During the 1990s, nearly all the language resources used for language processing were collections of texts. In the field of language processing, such collections are usually called corpora. Two of the best-known annotated corpora are corpora of English, the Brown Corpus and the Penn Treebank.

- The Brown Corpus (KUCERA; FRANCIS, 1982) includes a morphosyntactic annotation of words. It was compiled at Brown University and contains one million words.

- The Penn Treebank (MARCUS *et al.*, 1993) includes a morphosyntactic annotation of words (Fig. 1), but also a syntactic annotation of sentences of a part of the corpus, hence its name of 'treebank' (Fig. 2). It was compiled at the University of Pennsylvania and contains 4,5 million words.

*Battle-tested/JJ Japanese/JJ industrial/JJ managers/NNS here/RB always/RB buck/VBP up/RP nervous/JJ newcomers/NNS with/IN the/DT tale/NN of/IN the/DT first/JJ of/IN their/PP countrymen/NNS to/TO visit/VB Mexico/NNP ,/, a/DT boatload/NN of/IN samurai/FW warriors/NNS blown/VBN ashore/RB 375/CD years/NNS ago/RB ./.*

Fig. 1. Morphosyntactic annotation of the Penn Treebank. Each word is followed by a tag conveying morphosyntactic information.

```
(( (S
  (NP Battle-tested industrial managers
     here)
   always
  (VP buck
     up
     (NP nervous newcomers)
     (PP with
        (NP the tale
           (PP of
              (NP (NP the
                    (ADJP first
                        (PP of
                           (NP their countrymen)))
                    (S (NP *)
                      to
                      (VP visit
                         (NP Mexico))))
```

Fig. 2. Syntactic annotation of the Penn Treebank. Each phrase is enclosed in parentheses and begins with a tag conveying syntactic information.

### 1.2. Lexicons

The use of lexicons for language processing increased progressively in the recent years. Lexicons with only morphosyntactic information are now numerous, at least for inflectional languages. We will focus on more elaborate lexicons which also provide syntactic-semantic information. Examples of the most frequently cited lexicons for language processing are four lexicons of English: WordNet, FrameNet, VerbNet and ComLex.

- WordNet (MILLER, 1995) describe semantic relations. It was developed at Princeton University.

- FrameNet (FILLMORE; ATKINS, 1994), VerbNet (KIPPER *et al.*, 2000) and ComLex (GRISHMAN *et al.*, 1994) describe syntactic-semantic properties of verbs. They were respectively developed at Berkeley University, the University of Colorado at Boulder, and NewYork University.

### 1.3. Grammars

The use of grammars for language processing developed during the recent years, but less than that of lexicons. Each grammar conforms to a grammatical formalism, i.e. a way of expressing formal rules. Two examples of grammatical formalisms are tree adjoining grammars (TAG) and local grammars.

- The TAG formalism (JOSHI, 1985) is more powerful than the context-free grammar (CFG) formalism. It is used by several projects of construction of large-coverage grammars of English, French, Korean.

- The local-grammar formalism (GROSS, 1997) has the same expressive power as the CFG formalism. Local grammars are manually constructed, displayed and handled in a graphical, readable format. They are convenient for describing fine constraints involving lexical elements, as in Fig. 3.

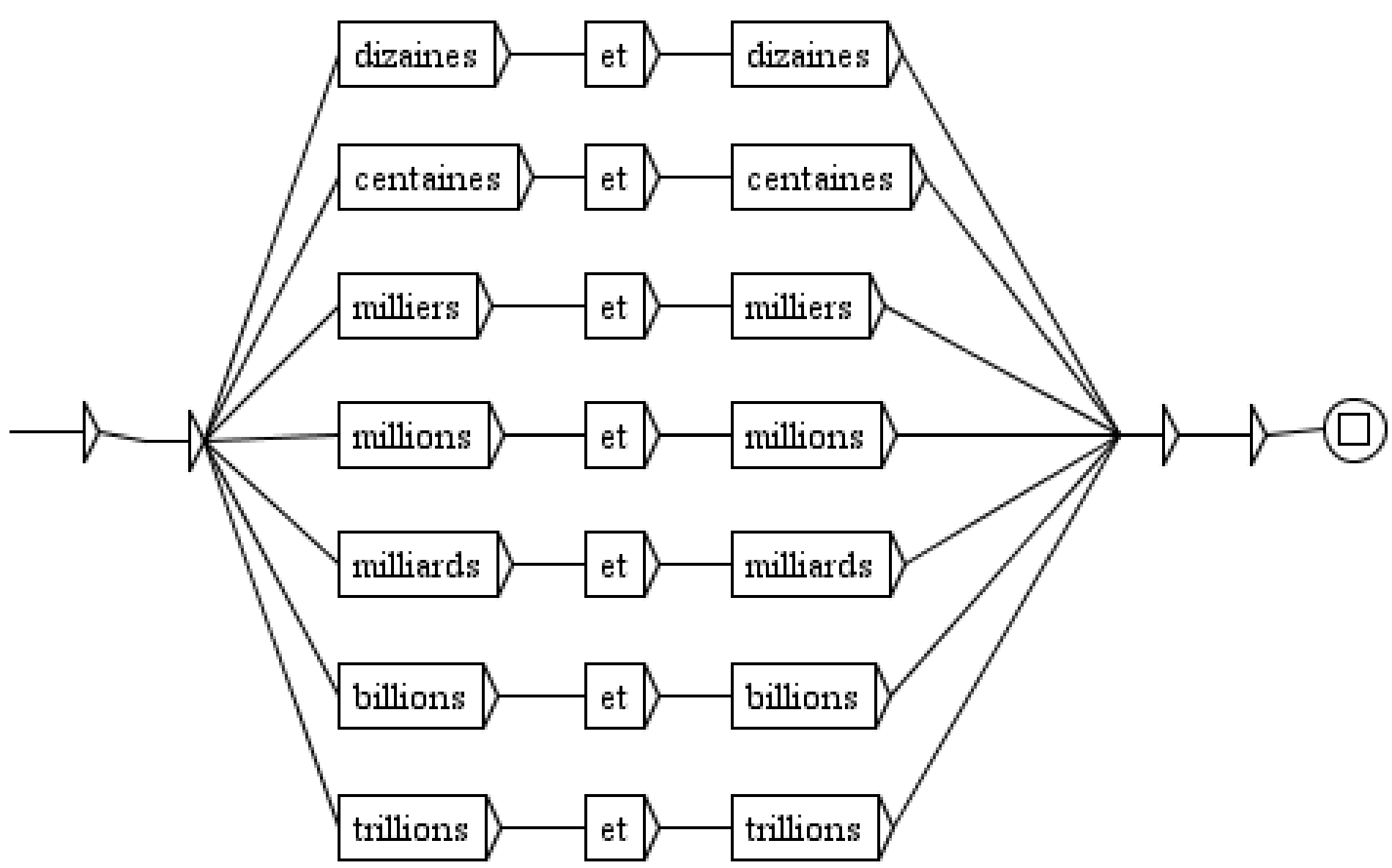

Fig. 3. Excerpt of a local grammar of French determiners.

## 2. Technical facts about manually constructed lexicons and grammars

Manually constructing a lexicon or a grammar for language processing requires expertise in linguistics, because it involves handling words and syntactic constructions, but also in information technology, since the results are to be exploited in computer applications. In this section, we outline a description of the activity of handcrafting lexicons and grammars.

### 2.1. Construction time

Most projects of lexicons for language processing require a lot of time and work.

It took 9 years (1998-2007) to complete 6100 FrameNet lexical entries of English such as the following two:

> [$_{Cook}$ *Matilde*] *fried* [$_{Food}$ *the catfish*] [$_{Heating_instrument}$ *in a heavy iron skillet*]
> [$_{Item}$ *Colgate's stock*] *rose* [$_{Difference}$ *$3.64*] [$_{Final_value}$ *to $49.94*]

The VerbNet project has almost the same rhythm with 5200 verbs in 9 years (1998-2007).

The ComLex project completed the description of 6000 verbs in 4 years (1993-1997), with for example the following formula for the verb *build*:

> *(verb :orth "build" :subc ((np) (np-for-np) (part-np :adval ("up"))))*

The Lexicon-Grammar project (GROSS, 1994), developed at University Paris 7, described 13000 entries of French verbs in 16 years (1968-1984), plus 10 000 nouns, 39 000 idioms and 10 000 adverbs, totalling 72 000 entries[3] in 30 years (1970-2000).

The Explanatory and Combinatorial dictionary (DiCo), developed at the Université de Montréal in the framework of the Meaning-Text theory (MEL'ČUK, 1981), completed 510 entries of French words in 23 years (1984-2007).

The Proton/Dicovalence project (EGGERMONT; VAN DEN EYNDE, 1990), developed at the Catholic University of Leuven, produced 8500 entries of French verbs in 6 years (1986-1992).

The Se-jong project (KIM, 2004), based at the National University of Seoul, achieved a remarkable work on Korean words in 9 years (1998-2007). The number of complete entries with syntactic-semantic description is not known to us.

Wordnet is an exception among lexicons. The first version, constructed in 5 years (1985-1990), comprised 70 000 synsets. A synset is a set of synonyms. Wordnet describes semantic relations, but not syntactic constructions.

The examples above show that lexicons with syntactic information require many years to reach sufficient coverage. In comparison, large annotated corpora are completed in a few years. The annotation of the Brown corpus (1 million words) took 9 years (1970-1979), but it was a pioneer. The annotation of the Penn Treebank took 3 years (1989-1992).

This contrast between the construction time of lexicons and of corpora is probably connected to the method of construction. Lexicons are usually handcrafted by teams of linguists, whereas the annotation of corpora is partially automated: the first version of the annotation is performed by programs.

### 2.2. Methodology

The construction of language resources involve specific methods, which have to do with applied linguistics.

Annotating a corpus consists essentially in observing forms occurring in a corpus that pre-exists to the study, and recognizing linguistic objects or linguistic phenomena in these forms. This process of observation and annotation is partially automatable: for example, annotation systems can assign parts of speech to words with a number of errors that depends on the method and on the language; the output is reviewed by linguists which correct errors. This was done on a large scale for the first time to the Brown corpus, with a tagging program (GREENE; RUBIN, 1971). The human work of observation and recognition is typical of corpus linguistics or corpus-based linguistics. We exemplify it by the use of a corpus in order to answer the following question[4]: do people say *abundant in* or *abundant with*? Using the English web as corpus, we can answer this question by observing that *abundant in* occurs 3,6 millions times whereas *abundant with* occurs 450 000 times. In order to refine this raw observation, we could analyse a selection of occurrences and check, for example:

- whether the preposition introduces a complement of *abundant*, as in *this region is abundant with wildlife*, or a complement of another linguistic element, as in *opportunities will be abundant with our help*;

- whether the content is denoted by the complement introduced by the preposition, as in *this region is abundant in wildlife*, or by the subject of *abundant*, as in *wildlife is abundant in this region*.

[3] More than half the entries are freely available on the web: http://infolingu.univ-mlv.fr/

Such a study would lead to revising the numbers of occurrences above, still relying exclusively on pre-existing texts.

As compared to this practice, the construction of lexicons and grammars involves the following operations, which are slightly more complex:
- fabricating examples methodically so that relevant parameters vary independently;
- submitting these forms to instrospective judgments of acceptability;
- deducing rules.

This process of manipulation is typical of introspective linguistics and requires human, active control. Let us exemplify it with the same question as above, i.e. *abundant in* vs. *abundant with*. We select a few possibly relevant parameters and we make them vary independently, as in the following fabricated examples (the asterisk * marks a sequence judged as unacceptable):

> *Dolphins are abundant in the bay*
> **Dolphins are abundant with the bay*
> *Dolphins are abundant around the bay*
> *The bay is abundant in dolphins*
> *The bay is abundant with dolphins*
> **The bay is abundant around dolphins*

Two parameters are manipulated in these sequences: the order of the arguments, *dolphins* and *the bay*, and the lexical value of the preposition: *in*, *with* or *around*. Such manipulation can be refined by checking (i) further values of the same parameters (e.g. *Dolphins are abundant off the bay*), and (ii) other parameters, e.g. lexical values of the arguments (*Calcium is abundant in broccoli*). The acceptabilities of the respective sequences leads to describing two constructions:

> *<Content> be abundant <Loc> <Place>*

where *<Loc>* is a locative preposition such as *in* or *around*, and

> *<Place> be abundant* (*in* + *with*) *<Content>*

where "+" symbolizes a choice between two lexical values, which are, in this case, free variants.

We consider such manipulation as the implementation of an experimental device. It is obviously a challenge to automation.

Both approaches, corpus linguistics and introspective linguistics, can be effective methods of observing the actual use of language, and they are compatible. However, the exclusive use of corpus linguistics would deprive the linguist of a valuable source of knowledge.

[4] This example is adapted from the introduction of Boons *et al.* (1976, pp. 34-46).

### 2.3. Information content

Language resources are used as stocks of information by language processing systems. Their quality depends much on the density of information that they contain: how much information? how precise? how exact? We will survey a few methods of assessing the information content of language resources.

2.3.1. Size of morphosyntactic tagsets

A simple measure of the density of morphosyntactic information is the size of the set of lexical tags used to represent this information. For example, the morphosyntactic information in the Brown Corpus is provided in the form of lexical tags which are assigned to words of the corpus. The size of the tagset is the number of different lexical tags. In the case of the Brown corpus, this number is between 87 and 119 tags, depending on sources. The morphosyntax of the Penn Treebank is expressed with only 36 tags. For instance, *milk* is tagged NN for 'noun, singular or mass', and *cars*, NNS for 'noun plural'.

Morphosyntactic tagsets of lexicons and grammars are usually more informative than those of annotated corpora. The Dela lexicons of French (COURTOIS, 1990) have a number of tags which can be reduced to 1000 tags without loss of information (LAPORTE; SILBERZTEIN, 1996). For instance, *lait* 'milk' is tagged [*lait* N m s], and *voitures* 'cars', [*voiture* N f p]. This tagset is larger mainly because tags contain the lexical value of lemmas, i.e. *lait* and *voiture*, in our example. In addition, tags possess a structure. They are made of several pieces of information, each of which is the value of a feature: in our example, lemma, part of speech, gender, and number. Structured tags used in lexicons and grammars can include information beyond morphosyntax.

2.3.2. Identification of constructions vs. identification of entries

Syntactic information is a major component of annotated corpora and lexicons. However, there exists a type of information which is present in some syntactic lexicons, but not in available annotated corpus. In order to present it, we will recall a technical distinction between identification of syntactic constructions and identification of lexical entries.

Identifying syntactic **constructions** means representing the corresponding syntactic structures and semantic properties. For example, the following sentences exemplify distinct constructions:

(1) *John charged the battery* *Nhum[agent] charge Nconc[theme]*
(2) *The battery charged* *Nconc[theme] charge*

In syntactic lexicons, syntactic constructions are identified through this type of formula, at least when they cannot be deduced from other information[5]. In syntactic grammars, syntactic constructions are also represented. In syntactically annotated corpora, differences between constructions are described in the annotations. For example, in Fig. 2, the construction of the verb *buck up* is identified by the presence of a subject, a direct complement, and an indirect complement (the modifier in *with*, interpreted with an instrumental meaning). Thus, the annotations identify the constructions.

The identification of lexical **entries** is slightly different. Sentences (1) and (2) above belong to one lexical entry, and sentence (3) to another:

(3) *The prosecutor charged Mark with fraud*
*Nhum[agent] charge Nhum[theme] with Nabs[event]*

[5] For instance, the verb *destroy* can enter in the same construction as (1): *John destroyed the battery*, but not as (2): **The battery destroyed*. This difference between *charge* and *destroy* is not known to be easily deduced from or clearly explained by other information about these verbs.

As a matter of fact, the semantic predicate found in (1) and (2) is not observed in (3), which has a completely unrelated meaning. The lexical entry for (3) can comprise other constructions, such as:

(4) *The prosecutor charged Mark*
*Mark was charged with fraud by the prosecutor*

The identification of lexical entries helps in establishing relations between words and meanings, in particular in case of ambiguous words, such as *charge*. Syntactic constructions are more a matter of how words are used in the context of sentences. Lexical entries and syntactic constructions are distinct notions, since the same meaning can be used in different constructions, e.g. (3) and (4), and the same construction with different meanings, e.g. (4) and (1).

Identifying lexical entries obviously falls within the scope of syntactic lexicons. The notion of lexical entry is a basic concept in lexicology. In our view, a syntactic grammar, or a syntactic annotation of a corpus, should also be able to connect constructions which convey the same meaning with the same word, such as (1) and (2), or (3) and (4). Such syntactic variations are examples of transformations (HARRIS, 1965) or alternations (LEVIN, 1993), a basic notion in syntax. Establishing such connections implies taking account of lexical entries, because of the preservation of meaning in the connected constructions.

In existing language resources, the distinction between identification of syntactic constructions and identification of lexical entries is often overlooked.

WordNet identifies entries, but not constructions. For example, (1) and (3) are described, but the syntactic information that (1) can also be expressed as (2), and (3) as (4), is missing.

ComLex identifies constructions, but not entries. For example, constructions such as (1), (2), (3) and (4) are described, but the semantic information that (1) and (2) belong to one entry and (3) and (4) to another is not systematically registered[6]. In the Penn Treebank and other available syntactically annotated corpora, such information is not registered at all.

Constructions and entries are identified as distinct objects in several other lexicons: the Lexicon-Grammar, DiCo, Dicovalence, FrameNet and VerbNet. It is not by chance that all these resources have been manually constructed.

2.3.3. Delimitation of frozen multi-word units

Frozen multi-word units, or compound words, are expressions which are typographically made of several words, but which linguistically require a description in a specific lexical entry, e.g. *keep in mind* or *hit the jackpot* (Fig. 4). Information about such words is fundamental. It is required for language-processing applications more elaborate than current systems (GROSS, 1986).

The delimitation of frozen multi-word units in most annotated corpora, e.g. the Brown corpus and the Penn Treebank, is very deficient. The French Treebank (ABEILLÉ *et al.*, 2003) is an exception: several categories of frozen multi-word units are systematically annotated as such. This annotation was derived from the Lexicon-Grammar, which is probably the most advanced lexicon in terms of description of frozen multi-word units.

2.3.4. Determinative nouns

A determinative noun is a noun that behaves as a determiner of another noun, as *part* in *They give back part of the loans*. In this sentence, if we take into account selectional restrictions, we will consider that the head of the object of *give back* is *loans*, rather than *part*, which quantifies *loans*. If we adopt a more morphosyntactic vision of syntax, we will analyse *part* as the head of the noun phrase. The latter analysis is systematically preferred in annotated corpora, e.g. in the Penn Treebank and in the French Treebank. Thus, in the Fench Treebank, *J'ai appris un certain nombre d'exigences administratives* 'I learned a number of administrative requirements' is analysed with *nombre* 'number' as the head of the object.

[6] The device provided by ComLex to encode such information is the frame-group clause, which in practice is not used to delimit complete lexical entries.

| N0 =: Nhum | N0 =: N-hum | Ppv | <ENT> | N0 V | <ENT>Det1 | N0 V N1 Prep N2 | <ENT>N1 | N1 =: Npc | [passif] |
|---|---|---|---|---|---|---|---|---|---|
| + | + | <E> | détendre | - | la | - | atmosphère | - | + |
| + | - | <E> | détenir | - | la | - | vérité | - | + |
| + | - | <E> | déterrer | - | la | - | hache de la guerre | - | + |
| + | + | <E> | détourner | - | la | - | conversation | - | + |
| + | - | <E> | détourner | - | la | - | tête | + | - |
| + | - | <E> | devancer | - | le | - | appel | - | + |
| + | - | se | dévisser | - | le | - | cou | + | - |
| + | + | <E> | dévorer | - | les | - | distances | - | - |
| + | + | <E> | dévorer | - | les | - | kilomètres | - | - |
| + | + | <E> | dévorer | - | la | - | route | - | - |

Fig. 4. Excerpt of a lexicon-grammar of French verbal idioms (GROSS, 1982).

## 3. Craftwork vs. industrial products

The preceding section showed that even the best annotated corpora, which are usually constructed semi-automatically in a few years, are less rich in fundamental linguistic information than lexicons and grammars, which are usually constructed manually by linguists, and require longer periods of time to be completed. All these differences are likely to be related. They evoke a contrast between an industrial process and a handcraft practice, and they suggest that the use of manually constructed language resources should be a factor of quality of language-processing applications.

Let us draw parallels in everyday life. Preparing an espresso coffee requires more time and manual work than an instant coffee, but the taste is more complex. The same can be said of home-made mayonnaise and commercial mayonnaise. In music, playing the violin requires more skill and effort than the electronic piano, but listeners find the sound more moving. In other words, it is trivially known that the industrialization of a product, although it is usually a progress, may lead to a loss of quality.

The handcraft approach and the industrial approach correspond to two current trends in language processing. Some specialists of the domain get progressively used to handling manually constructed resources, which are more informative and more complex, whereas others investigate in methods of automating and industrializing the construction of lexicons and grammars.

The trend towards manually constructed resources can be illustrated by the ongoing projects of syntactic and semantic lexicons (cf. sections 1.2. and 2). In addition, the community has engaged into projects of standardization of models of complex resources within the International Standardization Organization (ISO). Two standards are in preparation, the Morphosyntactic annotation framework or MAF (CLÉMENT; VILLEMONTE de LA CLERGERIE, 2005), exemplified in Fig. 5, and the Lexical markup framework or LMF (FRANCOPOULO *et al*., 2006).

The trend towards automation and industrialization of the construction of lexicons and grammars is illustrated by recent work on lexicon acquisition (SUN *et al*., 2008) and on probabilistic induction of grammars (KLEIN; MANNING, 2005). In these works, resources are automatically derived from corpora through statistical engineering. A few hot topics are: automatically classifying occurrences of words in order to discriminate their senses, e.g. charge in (1) and in (3); or automatically detecting which word combinations are frozen, like *keep in mind*.

```
<wordform entry='passagère' tokens='t1'>
    <fs>
      <f name='lemma'>
        <str>passager</str>
      </f>
      <f name='gramGrp'>
        <fs feats='pos@A gen@f num@s'/>
      </f>
    </fs>
  </wordform>
```

Fig. 5. Example of MAF-style encoding of morphosyntactic information about the French word *passagère* 'temporary'.

Choosing between these two options is not easy. The most desirable scenario in such situations is a fair competition, and then a choice determined by the respective success of each trend in making commercial applications possible, or according to other criteria. However, in the present case, these natural regulators do not work. None of the two trends shows sufficient maturity that it is about to be successfully exploited in real-size applications. None of them, thus, is expected to be validated through commercial computer applications any time soon. Current evaluation practices present glaring flaws:

- performances of systems are regularly assessed by reference to annotated corpora, but these corpora are not assessed themselves;

- tagsets are not evaluated either, not even compared;

- the possibility to improve systems once a dysfunction is identified is not estimated, not even mentioned as a desirable feature.

This uncertainty generates a tension between the two approaches. The Association for computational linguistics (ACL), one of the major actors of the field, is mainly in favour of the industrial approach: in the prestigious ACL-backed congresses, papers on handcrafted resources are rare, and when they are presented, the audience complains that the authors do not use statistics. The tension has also a corporative aspect: the handcrafting approach implies educating, training and hiring numerous linguists, whereas the industrial one depends essentially on computational and mathematical engineering.

This situation suggest a rivalry between linguists and computer scientists.

In our opinion, such a rivalry is quite absurd in the context of language processing. In fact, several types of interaction between linguists and computer scientists are essential to this domain. Linguistic expertise is required for the elaboration of formal models (cf. Fig. 5) which in turn are at the basis of the design of any computational approach. Linguistic expertise is also relevant to the construction and updating of language resources (cf. Fig. 4). Conversely, several computational aids to linguists' tasks require specific software. For example, the open-source, freely available Unitex system[7] (PAUMIER, 2006) contains:

- search tools, in particular a concordancer that generates lemmatized concordances of non-annotated corpora, facilitating the exploration of corpora,

- statistical tools that count the occurrences of each simple word,

- resource management tools, among which a generator of inflected forms, that facilitates the updating of the lexicons, and an editor of local grammars (cf. 1.3).

Thus, cross-interaction between linguists and computer scientists seems to us a condition of success for language processing. Instead of that, the question of whether language resources should be envisaged as craftwork or as industrial products is seldom seriously debated. Spärck-Jones (2007) contributes to this debate, but her article shows just how reluctantly each side is disposed to recognize the results of the other. This situation draws a methodological boundary in the middle of the field, hindering cooperation between actors that would benefit from working together. We would like to contribute to this debate by recalling and discussing a few scientific points.

## 4. Points in favour of industrialization

Authors in favour of the industrial approach to constructing language resources resort to three types of arguments : (i) manual construction of language resources is tedious and time-consuming; (ii) outcome contains subjective information; (iii) outcome is not formal enough for computational use.

### 4.1. Is manual construction of language resources tedious and time-consuming?

One of the most commonplace phrases in scientific articles about language processing is *tedious and time-consuming*. Authors use it to explain why their systems do not make use of manually constructed resources: their construction would be tedious and time-consuming. In scientific articles about language processing, according to the Google Scholar search engine, 7% of occurrences of *time-consuming* are closely associated with *tedious, laborious* or *labor-intensive*.

This argument shows a striking lack of scientific rigour.

Firstly, these researchers are right in assessing the cost of an approach before adopting it, and handcrafted language resources are admittedly costly in time, effort and skill; however, what about the quality of the outcome of the approach in question? These authors seldom combine the assessment of cost with an assessment of quality. Cost/quality ratio would be more relevant than only cost to making the methodological choice at stake. Overlooking the question of quality seems to imply that research should aim at minimizing costs, no matter the quality of the results to be expected. Avoiding hard work is surely a legitimate goal to a certain extent, but strategies to avoid hard work have been intensively investigated in 60 years of language processing, and the outcome comprises little reliable resources. When interesting results have been reached in the history of sciences in general, search for excellence was usually involved also.

Secondly, words like *tedious, laborious* or *boring* are an assessment of how much fun researchers find in their work. This is a question of personal taste, not a valid scientific point. Fun is certainly a part of a researcher's motivation for working, but only a part: the perspective of obtaining interesting results is also relevant, and is worth taking into account before giving up a tedious task. In addition, even though the average software engineer is not likely to find lexical description funny, other people are, including the author of this article. Idioms, for instance, are even a quite pleasant subject of work for a lexicon constructor. Now, usually, when you need something and do not feel like making it, you get it from other people, use it and assess it. If you do not like cooking, you will not necessarily fast. This is division of labour, one of the basic forms of economic exchange. If researchers find it tedious to manually construct language resources, why not experiment with resources produced by other researchers which are passionate on doing so?

[7] http://igm.univ-mlv.fr/~unitex

In our view, it is odd that the '*tedious and time-consuming*' argument is so widely used in the most prestigious scientific publications, in spite of the role of control and advice of peer reviewing. This is even shameful for the scientific committees of these publications.

### 4.2. Do manually constructed language resources contain objective information?

Manual construction of language resources is also frequently described as error-prone (*tedious and error-prone* is another Google-Scholar hit in language processing articles). In particular, when it is not exclusively based on corpora, but also on introspection, it is frequently criticized as subjective. Since introspective experimentation bears on the acceptability judgment of the experimentator, he/she is the object of his/her own experiment. This creates a risk of bias.

Three types of error are characteristic of a massive resort to introspection.

The first one stems from the linguist's insufficient ability to analyse sequences, and in particular to judge their acceptability. Even when linguists limit their work to their own native language, acceptability judgment is a skill which we do not all have to the same degree, just like almost any human activity.

The second risk is a difference between the language to be described and the descriptor's idiolect. We had a concrete example of this after one of our articles was reviewed: we had cited the French adverbial idiom *au petit bonheur la chance* 'haphazardly' in the form of *au petit bonheur de la chance*, the only one in our personal idiolect.

The third risk is that of an unconscious prejudice of the linguist, influenced by a desire to validate one of his/her hypotheses. For instance, there is a natural tendency to regularize phenomena. During the study of the nominalization relation established between the following sentences:

*Luc atterrit* 'Luc is landing' = *Luc fait un atterrissage* 'Luc is making a landing'

one may be temptated to overestimate the acceptability of sequence (5):

*Luc embraye* 'Luc is engaging the clutch'
= (5) *Luc fait un embrayage* 'Luc is engaging the clutch'

All these problems are well known to linguists that have a regular descriptive activity. They have an equivalent in any experimental science: the practical obstacles to the reproducibility of an experiment or of the measurement of a physical magnitude. An experiment, a measurement, have scientific relevance only if they are reproducible, that is, if other experimentators obtain the same results when they make it.

The requirement of reproducibility is as fundamental in linguistic description as in (other) experimental sciences. The three types of error listed above are three systematic causes of non-reproducibility of the observations and experiments needed for the construction of lexicons and grammars. They make it irrealistic to require absolute reproducibility, but if you analyse them, you draw the conclusion that they can be overcome in so far as there exists a linguistic community that speaks the language to be studied. As a matter of fact, if such a community exists, there is no reason why it would not produce speakers with various skills, including that of judging acceptability. The idiolect problem can be resolved in the same way, through a comparison of judgments emitted by various speakers. Finally, prejudices that can bias our judgments can be detected and fought with the aid of peer control.

In other words, the objectivity of the contents of manually constructed language resources depends on the practice of descriptors. Bad practices, i.e. lack of rigour and efficiency in reflecting actual usage of language, are stigmatised as 'armchair linguistics'. However, good practices and methodological provisions do exist. Provisions of a psychological nature, as in medical research, would be unrealistic. In medical research, such provisions consist in ensuring that the subject of an experimentation does not know what the outcome of the treatment he/she receives will mean for the results of the experimentation. In linguistics, equivalent provisions would mean something like ensuring that acceptability judgments are extracted from speakers who remain unaware of what the experiment consists of. The construction of a lexicon and a grammar of a language, which requires millions of acceptability judgments, would take thousands of years with such procedures. Methodological provisions of a linguistic nature are more relevant. We will mention three of them.

a) Observing examples attested in corpora is an invaluable aid to linguistic description. An efficient strategy consists in searching corpora for linguistic structures specified by their lexical and morphosyntactic content (lemmas, parts of speech, inflectional features), and generating the corresponding concordances[8], much more useful than those produced by concordancers without lexicon. Another strategy targets the web, and relies on common search engines or on the Webcorp system (RENOUF, 2003). Control through corpus observation has the advantage of involving large numbers of speakers. However, the exclusive use of this method does not ensure sufficient efficiency in reflecting actual usage of language. The reasons for this are well known and have been explained many times in discussions about the respective merits of introspective linguistics and corpus linguistics. Let us recall them briefly:

- Corpus observation does not provide analyses of meaning differences or of differences between variants of a language.

- It does not provide, by itself, a formalization of the facts observed.

- It does not attest inacceptabilities: for instance, the absence of the phrase *abonder de* in a French corpus of 820 000 words does not prove that this expression is not in use, and in fact, it is.

Thus, additional provisions are necessary. At a time when large corpora and corpus-processing tools were unavailable, the Lexicon-Grammar methodology equipped itself with an arsenal of corpus-independent methodological provisions, as a protection against risks related with fabricated examples and introspection.

b) The second provision consists in organizing regular meetings during which linguists control one another's judgments and analyses. Thus, the Lexicon-Grammar of French distributional verbs[9] (GROSS, 1975; BOONS *et al.*, 1976; GUILLET; LECLÈRE, 1992) was constructed during sessions with at least 5 linguists: Jean-Paul Boons, Jean Dubois, Maurice Gross, Alain Guillet and Christian Leclère, from 1969 to 1984. Presently, the Belgium-France-Quebec-Switzerland (BFQS) project on differences between verbal idioms in four variants of French (LABELLE, 1990; LAMIROY et al., 2003) involves meetings of 4 to 6 linguists (Fig. 6).

Indeed, collective sessions are needed to ensure that the description is accurate. Here is how the authors find the verbal idioms which are not used uniformly in the four variants of French. First, the representatives of each variant establish four separate lists. Then, they compare their lists. Realizing, for instance, that an idiom of list B (for Belgium) is not used in variant F (for France) requires a contact between a B representative and an F representative. This is because if an idiom of list B is not in list F, the author of the latter may simply have failed to notice it. In addition, if an idiom is in both lists B and F, that does not necessarily mean that it should be common to the two variants: in that case, interpretations must be compared; if they differ, there are, lexicologically, two idioms, each of which is in use in one of the variants and not in the other.

[8] For example, with the Unitex system.

[9] Distributional verbs are those which can be analysed as predicates. They are recognized by the fact that the distribution of arguments depends on the verb, as in *John plays dice*. They are opposed to support verbs (*John plays the king's part*) and verbal idioms (*John plays for time*).

| ***Idiom*** | B | F | Q | S | *Paraphrase* | *Exemple* |
|---|---|---|---|---|---|---|
| ***Amuser à des riens (s')*** | + | + | + | + | *Se distraire avec des futilités* | *Il est comme un petit enfant, il s'amuse à des riens.* |
| ***Amuser à un rien (s')*** | + | ! | – | + | *Se distraire avec des futilités* | |
| ***Amuser bien (s')*** | + | – | – | – | *Se plaire quelque part* | *Est-ce que tu t'amuses bien dans ton nouvel appartement ?* |
| ***Amuser la galerie*** | + | + | + | + | *Distraire l'assistance* | *"Lorsqu'il était petit, il amusait la galerie avec ses mimiques, ses blagues : un acteur était né. " (www)* |
| ***Amuser le tapis*** | – | + | – | + | *Distraire l'assistance* | *"Raffarin veut-il amuser le tapis ? Après tout, pourquoi pas, mais la situation dramatique de la France mérite mieux." (www)* |
| ***Amuser le temps*** | – | – | + | – | *Faire passer le temps* | *Pierre n'a rien fait de la journée. De plus en plus, j'ai l'impression qu'il amuse le temps.* |

Fig. 6. Sample of the BFQS lexicon of verbal idioms.

The main drawback of the practice of collective sessions is its cost.

c) The third provision consists in systematically scrutinizing the criteria of verification of the syntactic-semantic features under study, and assessing the reproducibility of the application of these criteria. For example, some transitive distributional verbs admit the neutrality transformation (BOONS *et al.*, 1976) or causative alternation (LEVIN, 1993):

(6) *John breaks the branch* = (7) *The branch breaks*

This property might seem simple, but studying it on a couple of examples is one thing, and encoding it consistently over the whole lexicon of verbs is a harder task. One of the criteria retained to determine if a verb admits this syntactic feature consists in applying the transformation formally, i.e. according to the $N_0\ V\ N_1\ W = N_1\ V\ W$ formula, which is a representation of (6) = (7), and judging the acceptability of the result:

*John watches the landscape* **The landscape watches*

This is a formal criterion. Experience shows that formal criteria are usually of a much higher reproducibility than most other types of applicable criteria. In order to benefit from this effect, the syntactic-semantic features represented in lexicon-grammars are based, as often as possible, on formal criteria set out in detail in the books, articles and theses published with the lexicons.

Some properties cannot be characterized through formal criteria only and require to take into account semantic criteria. In that case, differential semantic evaluation is preferred to absolute semantic evaluation, as being more reliable (GROSS, 1975). A characterization of how (6) and (7) differ semantically will probably be more reproducible than an absolute characterization of what (6) means. Now if we compare the (6)-(7) semantic difference with (8)-(9):

(8) *John weighs the bag* (9) *The bag weighs*

the characterization will probably be even more reproducible. The (6)-(7) difference, which has to do with the cause of the process, is not encountered at all in (8)-(9). This concept of differential semantic evaluation lies at the heart of Z. Harris' notion of transformation: the $N_0\ V\ N_1\ W = N_1\ V\ W$ formula will be considered a transformation only if the (6)-(7) semantic difference is encountered in a sufficient number of pairs with the same structures and other lexical material.

Handcrafting language resources is often criticized as being a subjective activity. However, this criticism is seldom supported by a discussion of the methodological provisions described above, or by an actual analysis of resulting resources. This makes this criticism little convincing.

### 4.3. Are manually constructed language resources formal?

The result of a task of linguistic description may be usable or not for language processing. For example, in usual dictionaries, most information is provided in the form of text: examples, definitions, comments... and is not ready to use in computer programs (GROSS, 1989; IDE; VÉRONIS, 1993). Information technologies require that resources to be exploited by programs conform to a formal model. In a formal model, information elements are unambiguous and are placed in explicit structures, such as tables, trees, graphs... This is not the case of information conveyed by text. Access to such information implies interpreting the text, which always contains ambiguous parts; the structure of the information is not explicit, but is understood by the reader. On the contrary, any computer program requires a formal model. The model abstracts away whatever is of no interest, making up a simplified version of the reality that can be handled by the program.

Whether the outcome of a task of linguistic description is formal enough to be usable for language processing depends all on the practice of the descriptor. In this section, we present a few notions that are useful bases for formal models of language resources.

One of these fundamental notions is that of lexical entry. Describing the noun *bank* implies distinguishing at least two entries, one meaning a financial institution, the other the edge of a river. In this case, the two entries have quite different etymological histories, but even when two senses of the same word share much of their etymological history, this closeness is usually of no help for language processing. For example, consider two senses of the verb *float*:

*The oil floats on the water*
*The euro floats against the dollar*

The syntactic constraints satisfied by the verb are different in the two sentences, the translation into another language may be different... Various types of information to be attached to each lexical entry are essential to meaning-related applications; however, knowledge about metaphorical, historical or cognitive relations between the two entries, even if it plays a prominent role in our intuitions about these senses, is unlikely to be exploitable. This is why the notion of lexical entry is fundamental in formal models underlying language resources. So is the notion of syntactic construction (cf. 2.3.2, examples (1), (2) and (3)).

In the formal model of a lexicon, lexical entries are objects which are assigned features such as syntactic constructions and semantic features. It is important for the consistency of the model that each feature be applicable to each entry, i.e., it must be possible to decide the value of each feature for each entry. This is not so easy to obtain with semantic features. For example, if we feel that the French preposition *devant* 'in front of' conveys a notion of domination, integrating such a feature into a formal resource implies defining it or finding a criterion for it, so one can tell the lexical entries that have this feature from those that do not. Some features turn out to be formalizable, others do not.

This formalization work may seem new to some linguists. It is not only a matter of observing examples in a corpus or elsewhere: it also means active manipulation by speakers, formulation and testing of hypotheses etc. The industrial approach to language resources is an attempt to automate such activity.

More and more manually constructed language resources are available, and most of them have a satisfying degree of formalization. Thus, the idea that handcrafting language resources is incompatible with formalization is more an overgeneralization about linguists' work than a serious criticism of particular projects or resources.

We surveyed three points in favour of the industrial approach. After analysis, it appears that these points are more researchers' personal reasons to avoid the craftwork approach than convincing evidence that this approach would be unrealistic or would give results of insufficient quality.

## 5. Points in favour of the craftwork approach

The craftwork approach takes advantage of human skills in order to produce formal rules and data. For example, the method described in section 2.2. to investigate into the syntax of *abundant* requires insight, ability to create hypotheses, and ability to check them. A reason to adopt the craftwork approach is that it is dubious that these three human skills might be simulated by computer programs beyond very simple cases. The reasoning that underlies the example of section 2.2. involves identifying relevant parameters: the order of the arguments (*dolphins are abundant in the bay* vs. *the bay is abundant in dolphins*), the lexical value of the preposition (*in*, *with*, *around* or others), the lexical content of the arguments, the determiners etc. Then, examples are devised so that each parameter varies independently. As in any experimental science, the manipulator designs experiments in order to examine separately the effects related to the different parameters that could be factors of the phenomena that have been observed. This leads to validating the respective hypotheses underlying these experiments, or to imagining other hypotheses. Even if a corpus is used as a stock of examples, the discovery and formalization of rules goes far beyond the observation of examples: it involves a much more active and methodical exploration of the possibilities of the language.

A reason why the industrialization approach is adventurous is that the only applicable method of inferring rules from examples is statistic generalization, which is technically incompatible with complex structures, i.e. structures with numerous parameters, each of which can take numerous values. The space and time required to explore the possible rules grows exponentially, and therefore very fast, with the complexity of the objects handled by programs of automatic generalization. Lexicons and grammars represent syntactic objects: lexical entries, syntactic constructions, contexts..., which are complex structures.

There is also a technical reason to adopt the craftwork approach. When a system is based on manually constructed resources, and when a dysfunction is identified, it is usually possible to understand the cause of the problem and to directly modify the resource so that the operation of the system improves. (Manually constructed resources are often readable and updatable, precisely because manual construction requires that they should be.) In contrast, systems obtained through statistic generalization can only be modified by training them further with further corpus. When a dysfonction is identified, further training is not sure to correct or even alleviate it, and may deteriorate other aspects of the operation of the system.

## Conclusion

Schematically, there exist two approaches to the construction of language resources: direct manual construction by linguists, which we called the handcrafting approach, and automated construction with the aid of statistic-based computer programs, which we called the industrial approach. This methodological opposition is seldom debated. We recalled technical facts which suggest that the handcrafting approach should be a factor or quality of language-processing applications. We discussed some scientific arguments presented on both sides. It appears that researchers' reasons to avoid the craftwork approach are more personal preferences than convincing evidence that this approach would be unrealistic or inefficient.